\documentclass[sigconf]{acmart}

\usepackage{multirow}
\usepackage{pifont}
\AtBeginDocument{%
  }

\setcopyright{acmlicensed}
\copyrightyear{2025}
\acmYear{2025}
\acmConference[MM '25] {Proceedings of the 33rd ACM International Conference on Multimedia}{October 27--31, 2025}{Dublin, Ireland.}
\acmBooktitle{Proceedings of the 33rd ACM International Conference on Multimedia (MM '25), October 27--31, 2025, Dublin, Ireland}
\acmISBN{979-8-4007-2035-2/2025/10}
\acmDOI{10.1145/3746027.3755487}

\begin{document}

\title{PRINTER:Deformation-Aware Adversarial Learning for Virtual IHC Staining with In Situ Fidelity}

\author{Yizhe Yuan}
\authornote{Both authors contributed equally to this research.}
\orcid{0009-0007-8602-6322}

\affiliation{%
  \institution{Shanghai Jiao Tong University}
  \city{Shanghai}
  \country{China}
}
\email{yuanyz0825@sjtu.edu.cn}

\author{Bingsen Xue}
\authornotemark[1]
\orcid{0000-0002-6503-4741}
\affiliation{%
  \institution{Shanghai Jiao Tong University}
  \city{Shanghai}
  \country{China}
}
\email{bingsenxue@sjtu.edu.cn}

\author{Bangzheng Pu}
\orcid{0000-0003-1097-6819}
\affiliation{%
  \institution{Shanghai Jiao Tong University}
  \city{Shanghai}
  \country{China}}
\email{bangzheng@sjtu.edu.cn}

\author{Chengxiang Wang}
\orcid{0009-0000-9498-7587}
\affiliation{%
 \institution{Shanghai Jiao Tong University}
 \city{Shanghai}
 \country{China}}
\email{wcx123@sjtu.edu.cn}

\author{Cheng Jin}
\authornote{Corresponding author.}
\orcid{0009-0003-7579-0035}
\affiliation{%
  \institution{Shanghai Jiao Tong University}
  \city{Shanghai}
  \state{China}
  \country{chengjin520@sjtu.edu.cn}}
\renewcommand{\shortauthors}{et al.}
\begin{abstract}
Tumor spatial heterogeneity analysis requires precise correlation between Hematoxylin and Eosin (H\&E) morphology and immunohistochemical (IHC) biomarker expression, yet current methods suffer from spatial misalignment in consecutive sections, severely compromising in situ pathological interpretation. In order to obtain a more accurate virtual staining pattern, We propose PRINTER, a weakly-supervised framework that integrates \textbf{PR}ototype-dr\textbf{I}ven content and stai\textbf{N}ing pat\textbf{TER}n decoupling and deformation-aware adversarial learning strategies designed to accurately learn IHC staining patterns while preserving H\&E staining details. Our approach introduces three key innovations: (1) A prototype-driven staining pattern transfer with explicit content-style decoupling; and (2) A cyclic registration-synthesis framework GapBridge that bridges H\&E and IHC domains through deformable structural alignment, where registered features guide cross-modal style transfer while synthesized outputs iteratively refine the registration;(3) Deformation-Aware Adversarial Learning: We propose a training framework where a generator and deformation-aware registration network jointly adversarially optimize a style-focused discriminator. Extensive experiments demonstrate that PRINTER effectively achieves superior performance in preserving H\&E staining details and virtual staining fidelity, outperforming state-of-the-art methods. Our work provides a robust and scalable solution for virtual staining, advancing the field of computational pathology.
\end{abstract}

\ccsdesc[500]{Applied computing~Life and medical sciences}

\keywords{Virtual Staining, Weakly-Supervised, Prototype-Driven, Deformation-Aware Adversarial Learning, Computational Pathology}


\maketitle
\begin{figure}[t] 
  \centering
  \includegraphics[width=\linewidth]{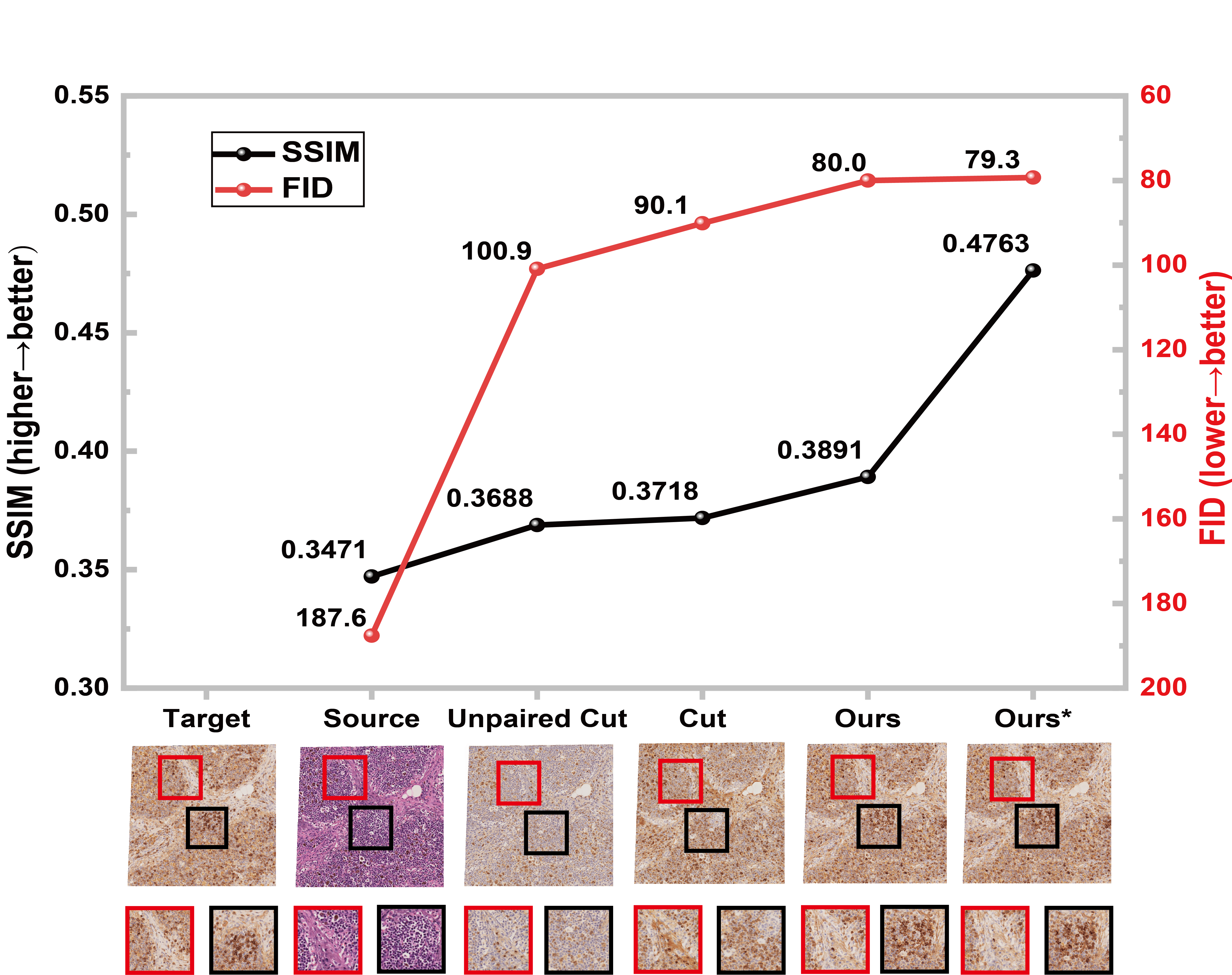} 
  \caption{Structural Misalignment Hinders Domain Adaptation: An SSIM and FID Analysis. From left to right: original H\&E and IHC; image generated by unpaired training setup CUT; image generated by paired training setup CUT; image generated by our method; and the image after applying the deformation.}
  \label{Fugure1}
\end{figure}
\section{Introduction}
Understanding the spatial heterogeneity of tumors and its association with disease initiation and progression is a cornerstone of cancer biology. Current histopathological workflows heavily rely on Hematoxylin and Eosin (H\&E) staining and immunostaining of serial sections through immunohistochemistry (IHC) staining \cite{pati2024accelerating}. However, this process is cumbersome, tissue-exhaustive, and prone to misalignment of tissue images. Additionally, compared to H\&E staining, IHC staining involves a more complex and expensive preparation process, limiting its widespread application in clinical practice \cite{li2024virtual}. Therefore, translating H\&E to IHC staining could serve as an ideal solution for obtaining IHC-stained images.  \par
Recent advances in GANs have significantly enhanced virtual staining \cite{cyclegan, cut, liu2024virtual}. However, current methods are typically constrained to either fully unpaired training (with \cite{pati2024accelerating} or without \cite{yan2023unpaired} manual annotations) or strictly \cite{de2021deep} aligned paired data. Since it is often hard to acquire strictly aligned paired data, recent work has proposed methods using loosely \cite{asp, psp, ppt} aligned pairs to extract optimal supervision, providing a more adaptable and practical solution. However, existing methods either tolerate misalignment or extract supervised signals from neighboring slices, but challenges of structural inconsistencies still remain, including preserving the staining details of the original H\&E images and learning the correct IHC staining. \par
The difficulty of IHC virtual staining from  serial H\&E sections lies in the following aspects: 1) \textit{The specificity of IHC staining}.
IHC staining is designed to detect the expression of specific target proteins (e.g., CD3, PAX5, ER, HER2), for specific cells or tissue regions, while H\&E staining provides comprehensive histomorphological information. 2) \textit{Structural inconsistencies caused by serial sectioning}. Although serial sections from the same specimen typically have a thickness of only 2--5 \textmu m, minor spatial displacements and morphological variations between adjacent sections may still occur during sections preparation. 3) \textit{Conflict between content preservation and staining pattern learning}. The virtual staining model must maximally preserve the original tissue morphology and detailed information from HE images. This requires the model to capture the key signal characteristics of IHC during cross-domain mapping without neglecting the holistic structural information provided by H\&E staining. \par
The essence of virtual staining is image transformation. To tackle structural inconsistency, besides accepting pixel-level differences between adjacent slices or using unpaired training, another approach is to reduce these inconsistencies with image alignment techniques \cite{li2024virtual}. One common strategy aligns the generated image—after it passes through the discriminator—with the target using L1/L2 loss \cite{mae}. While this improves spatial alignment, it often fails to preserve fine-grained staining details and leads to discrepancies in data distribution \cite{grit, dfmir}. In fact, there is an essential problem, which is that structural inconsistency can lead to differences in data distribution\cite{honkamaa2023deformation, tpami}. In addition, keeping the generated modality content-consistent with the original modality \cite{pati2024accelerating, cut, jiang2023scenimefy}, can lead to conflicts in the learning of the staining patterns. A better strategy is to decouple the content and the staining pattern, typically employing learnable encodings to drive staining modality learning \cite{zhang2022mvfstain, guan2024unsupervised}. However, the intricacy of staining patterns poses a significant challenge in designing a semantically rich staining framework to guide staining synthesis. \par
To address these challenges, we developed a novel weakly supervised intensity and structure unbiased style transfer model, named PRINTER, for high-quality H\&E to IHC virtual staining generation. Specifically, we introduced an explicit content-style decoupling method based on learnable prototypes, which successfully resolved the fundamental conflict between content preservation and style migration in cross-modal virtual coloring by decomposing IHC coloring patterns into a dictionary of migratable style prototypes. Subsequently, we established a cyclic alignment synthesis framework, GapBridge, which achieved accurate cross-domain alignment between the generated and target images by iteratively optimizing deformable alignment and virtual staining. Additionally, we aimed to eliminate the interference of structural discrepancies on the discriminator, enabling it to focus on distinguishing staining patterns. We provided a structurally distorted throw pressure to the generator through the alignment network, jointly deceiving the discriminator. Meanwhile, the discriminator was designed to refine coloring details as closely as possible to accurately distinguish structurally consistent images. Fig. \ref{Fugure1} illustrated the learning gain in staining patterns from unpaired to paired training CUT \cite{cut}, as well as the gradual reduction of structural inconsistency using our method. \par
The primary contributions of this work are three-fold:

\begin{enumerate}
    \item \textbf{Prototype-Driven Virtual Staining}: We propose a novel prototype-driven framework for precise virtual staining from H\&E to IHC images, which effectively resolves the fundamental conflict between content preservation and staining pattern learning. 
    \item \textbf{Dual-Network Adversarial Learning}: We introduce an innovative cooperative adversarial framework where a jointly trained generator and registration network collaboratively challenge the discriminator. This unique architecture forces the discriminator to develop more discriminative capabilities at both tissue and cellular levels, serving as a \textit{GapBridge} that significantly enhances the quality of virtual staining generation.

    \item \textbf{Comprehensive Experimental Validation}: Through extensive quantitative and qualitative experiments, we demonstrate the superior performance of our method in preserving H\&E structural details while achieving precise stain pattern transfer. Ablation studies provide insights into virtual staining mechanisms, offering valuable perspectives for future research in computational histopathology.
\end{enumerate}
\begin{figure*}[t] 
  \centering
  \includegraphics[width=\linewidth]{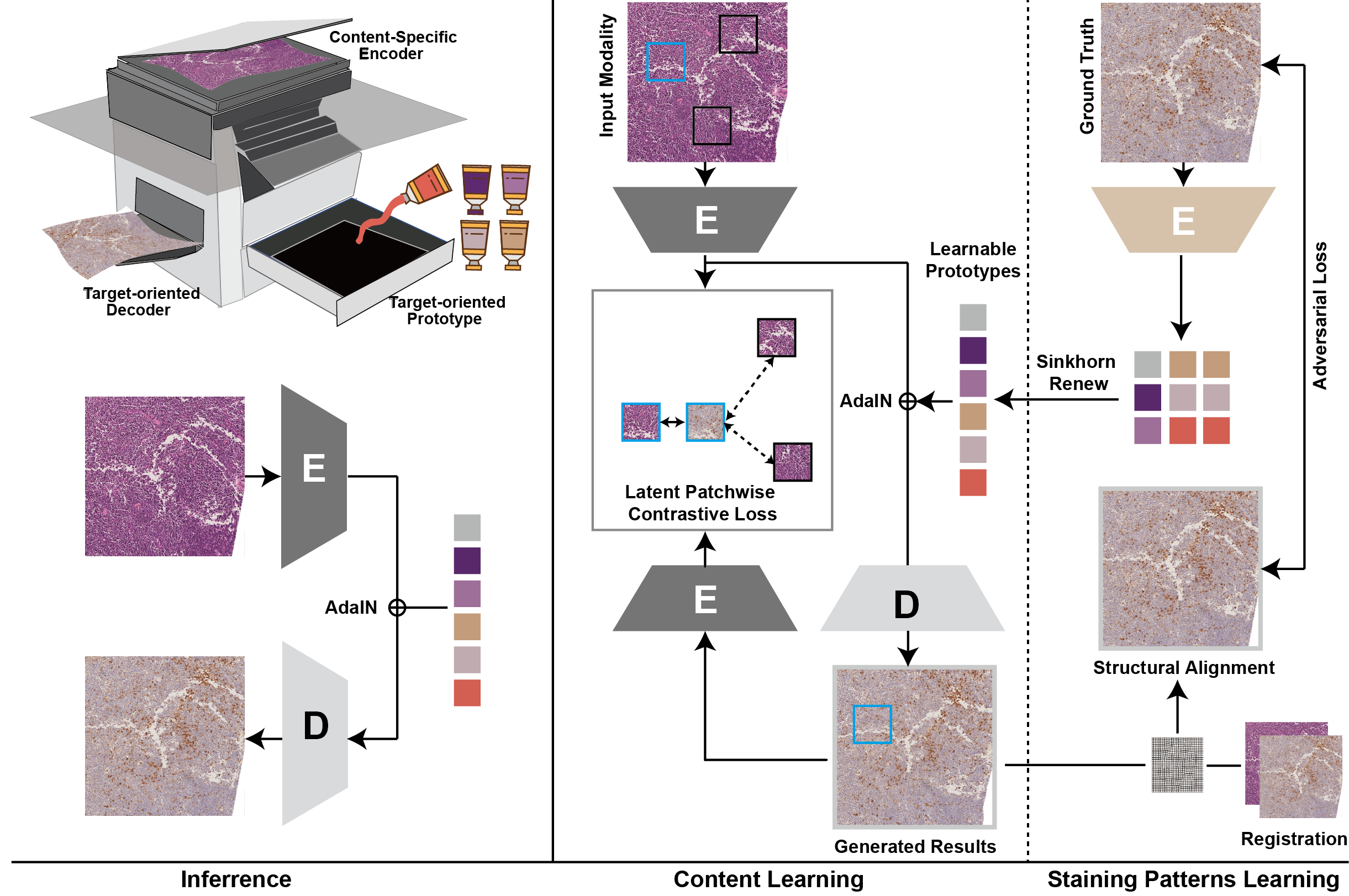} 
\caption{
LEFT: H\&E content features extracted by encoder are adaptively fused with learned prototypes via AdaIN to generate virtual staining results. 
MIDDLE: Content learning phase: generating image and target image content consistency in latent space via Patchwise Contrastive Loss constraints. 
RIGHT: Staining patterns learning: A structurally consistent image joint spoofing discriminator is generated by an alignment network and accurately learned staining patterns are migrated to unaligned images.
}
  \label{fig:teaser}
\end{figure*}

\begin{figure*}[t]
    \centering
    \includegraphics[width=\linewidth]{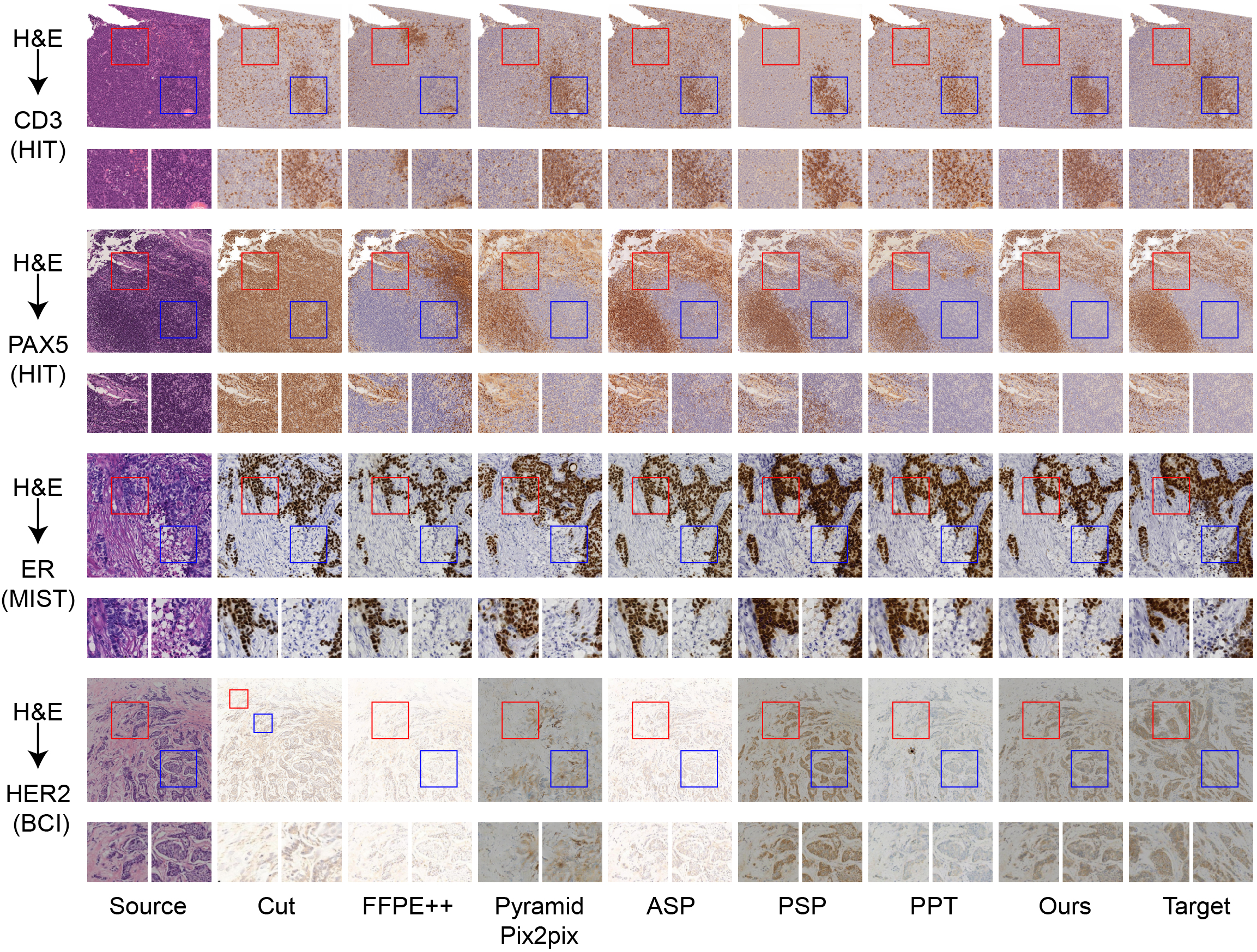}
    \caption{Qualitative comparison of virtual staining results from H\&E to four immunohistochemical markers: CD3, PAX5, ER, and HER2.}
    \label{fig:Compare_result} 
\end{figure*}
\section{Related Work}
\subsection{Image-to-image translation}
Image translation techniques aim to maintain the content of the original image while mimicking the stylistic features of the target domain \cite{li2024virtual1,src,wu2024stegogan}, which are mainly categorized into paired \cite{yu2024unpaired} and unpaired methods. Pix2Pix \cite{pixel2pixel} pioneered pixel-level supervised transformation based on the conditional GAN \cite{li2023bbdm,jiang2023scenimefy} for pixel-level supervised transformation, and Pix2PixHD \cite{wang2018high} and InstructPix2Pix \cite{brooks2023instructpix2pix} respectively improve the resolution and command control capabilities, respectively. To address the data matching challenge, CycleGAN \cite{cyclegan} adopts the cyclic consistency assumption, but this bijection assumption often does not hold in structured transformations. CUT \cite{cut} breaks through this limitation and enhances the local feature alignment through contrast learning, which demonstrates the advantage in stylized transformation \cite{ pati2024accelerating}. Existing methods, although suitable for virtual staining from H\&E to IHC, do not fully utilize the inherent near-pairing property of continuous slices.

\subsection{Virtual Staining in Digital Pathology}
The earliest virtual staining models operated in a paired setting using various types of GANs \cite{wolflein2023hoechstgan,liu2023vsgd}. Successful cases of paired models include translating H\&E images to special stains such as PAS, PAM, and MASSON \cite{de2021deep,yang2022virtual}, as well as translating H\&E to IHC \cite{liu2022bci}. However, in practice, re-staining is not routinely practiced, and sequential sections are difficult to align accurately. For this reason, the unpaired staining conversion model was developed for H\&E to IHC specialized staining conversion and frozen-FFPE section conversion\cite{ozyoruk2022deep,kassab2024ffpe++}. While these unpaired translation methods alleviate the need for perfect alignment, they often sacrifice staining accuracy due to a complete lack of structural guidance. The ASP method treats neighboring slices as noisy paired positive samples and tolerates pixel-level misalignment by dynamically suppressing mismatched regions through an adaptive weighting mechanism. The PPT method introduces the patch alignment method as a weakly aligned reference datum for staining transitions \cite{ppt}. The PSP method propagates prototypes of protein expression from sparsely annotated neighboring slices to guide molecular sensing staining transitions \cite{psp}. However, these methods, while partially solving the problem of misalignment between neighboring slices, fail to radically reduce the differences in the distribution of structural domains caused by structural inconsistencies.
\subsection{Image Registration and Translation}
Deformable medical image registration using deep learning has gained significant attention recently \cite{wang2024recursive,khor2023anatomically,meng2024correlation}. Traditional cross-modal image registration methods typically align raw images acquired from different modalities (e.g., CT, MRI, PET) \cite{mok2024modality, ghahremani2024h,dong2023preserving}. However, due to substantial differences in imaging principles and feature representations across modalities, direct cross-modal registration remains highly challenging \cite{zhao2025deformation}. To mitigate this, recent studies have proposed cross-modal image translation methods \cite{dfmir,honkamaa2023deformation}, which reduce inter-modality discrepancies by translating source modality images into pseudo-target modality images. Nevertheless, existing works frequently overlook the potential of using registration to bridge domain gaps for better image translation.

\begin{table*}[htbp]
    \centering
    \caption{Evaluation Metrics for Different Datasets and Methods}
    \label{tab:evaluation_metrics}
    \begin{tabular}{@{}lccccccc|c@{}}
        \toprule
        Dataset & Method & \multicolumn{2}{c}{Content Consistency} & \multicolumn{4}{c}{Staining pattern Consistency} & Avg. Rank \\
        \cmidrule(lr){3-4} \cmidrule(lr){5-8}
        & & SSIM↑ & PSNR↑ & SSIM↑ & PSNR↑ & LPIPS↓ & FID↓ & ↓\\
        \midrule
        
        \multirow{7}{*}{H\&E $\to$ CD3 (HIT)} 
          & CUT \cite{cut} & {\color{blue}0.7600} & {\color{blue}17.4210} & 0.3718 & 17.8735 & 0.2815 & 90.0959 & 3.3\\
          & FFPE++7 \cite{kassab2024ffpe++} & 0.6810 & 16.3589 & 0.3728 & 17.4980 & 0.3040 & 95.1369 & 5.3\\
          & PyramidPix2Pix \cite{liu2022bci} & 0.3535 & 14.6172 & 0.3699 & {\color{blue}18.2900} & 0.2814 & 101.6701 & 5.2\\
          & ASP \cite{asp} & 0.7180 & 16.9185 & 0.3655 & 17.5943 & 0.2846 & 87.7009 & 4.7\\
          & PSP \cite{psp} & 0.7297 & 15.3891 & {\color{blue}0.3769} & 17.7803 & {\color{blue}0.2675} & {\color{blue}85.3342} & 3.2\\
          & PPT \cite{ppt} & 0.6501 & 16.4615 & 0.3694 & 17.4713 & 0.2913 & 86.1170 & 5.3\\
          & Ours & {\color{red}0.7926} & {\color{red}17.8303} & {\color{red}0.3891} & {\color{red}18.5368} & {\color{red}0.2501} & {\color{red}80.0493} &1.0\\
        \midrule
        
        \multirow{7}{*}{H\&E $\to$ PAX5 (HIT)}
          & CUT \cite{cut} & 0.6239 & 15.6086 & 0.3136 & 16.5285 & 0.3372 & 94.4340 & 5.2\\
          & FFPE++ \cite{kassab2024ffpe++} & 0.6360 & 15.2646 & 0.3133 & 16.5407 & 0.3412 & 98.4873 & 5.8\\
          & PyramidPix2Pix \cite{liu2022bci} & 0.2906 & 13.7026 & 0.3063 & 17.1182 & 0.3140 & 94.5621 & 6.2\\
          & ASP \cite{asp} & 0.6778 & {\color{blue}16.0870} & 0.3164 & 17.1589 & 0.3041 & 85.1021 & 3.2\\
          & PSP \cite{psp} & {\color{blue}0.7008} & 14.4579 & {\color{red}0.3320} & {\color{red}17.4880} & {\color{red}0.2892} & 85.1455 & 2.5\\
          & PPT \cite{ppt} & 0.6098 & 15.5375 & {\color{blue}0.3215} & {\color{blue}17.4177} & 0.3006 & {\color{blue}84.6294} & 3.2\\
          & Ours & {\color{red}0.7161} & {\color{red}16.1967} & 0.3146 & 17.2136 & {\color{blue}0.2936} & {\color{red}81.4670} & 2.0\\
        \midrule
        
        \multirow{7}{*}{H\&E $\to$ ER (MIST)}
          & CUT \cite{cut} & 0.7022 & {\color{red}13.9971} & 0.1781 & 13.5509 & 0.4551 & 38.9496 & 4.3\\
          & FFPE++ \cite{kassab2024ffpe++} & {\color{blue}0.7133} & 13.6216 & {\color{blue}0.1822} & 13.7272 & 0.4645 & 38.7274 & 4.2\\
          & PyramidPix2Pix \cite{liu2022bci} & 0.1282 & 11.2563 & 0.1768 & {\color{blue}14.8679} & 0.4484 & 70.6640 & 5.5\\
          & ASP \cite{asp} & {\color{red}0.7304} & {\color{blue}13.6786} & 0.1773 & 14.3131 & 0.4332 & {\color{blue}33.6365} & 3.0\\
          & PSP \cite{psp} & 0.6742 & 12.8929 & 0.1573 & 14.3406 & 0.4307 & 35.9693 & 4.7 \\
          & PPT \cite{ppt} & 0.6337 & 13.5170 & {\color{red}0.1828} & 14.6017 & {\color{red}0.4166} & 33.9854 & 3.0\\
          & Ours & 0.6708 & 13.3091 & 0.1675 & {\color{red}14.9855} & {\color{blue}0.4287} & {\color{red}32.0808} & 3.3\\
        \midrule
        
        \multirow{7}{*}{H\&E $\to$ HER2 (BCI)}
          & CUT \cite{cut} & {\color{blue}0.7074} & 9.9543 & 0.3694 & 14.7774 & 0.4587 & 51.1016 & 4.3\\
          & FFPE++ \cite{kassab2024ffpe++} & 0.6872 & 9.5321 & {\color{blue}0.4082} & 14.7326 & 0.4507 & 54.1656 & 4.8\\
          & PyramidPix2Pix \cite{liu2022bci} & 0.2878 & {\color{blue}16.5655} & 0.3593 & 16.6875 & 0.4962 & 125.8636 & 5.7\\
          & ASP \cite{asp} & 0.7035 & 9.8983 & 0.3663 & 14.9383 & 0.4420 & 68.0847 & 4.8\\
          & PSP \cite{psp} & 0.6640 & 14.6418 & 0.3951 & {\color{red}19.5338} & {\color{red}0.4269} & {\color{red}38.7137} & 2.7\\
          & PPT \cite{ppt} & 0.6709 & 11.7057 & 0.3646 & 16.8532 & {\color{blue}0.4250} & 62.3313 & 4.0\\
          & Ours & {\color{red}0.7315} & {\color{red}18.1639} & {\color{red}0.4368} & {\color{blue}19.1816} & 0.4271 & {\color{blue}42.1208} & 1.7\\
        \bottomrule
    \end{tabular}
\end{table*}

\section{Methodology}
The virtual staining process can be imagined as a copy printer, which preserves tissue morphology from input sections and renders outputs by learned staining patterns. Therefore,  PRototype-drIven content
and staiNing patTERn decoupling and deformation-aware adversarial learning strategies (PRINTER) are proposed to resolve the inherent conflict between staining pattern fidelity and content preservation, as well as the challenges in achieving precise staining pattern transfer. 


\begin{table}[t]
\centering
\caption{Content Consistency Evaluation ($X$ vs $\hat{Y}$)}
\label{tab:staining_consistency}
\begin{tabular}{lcc}
\toprule
Method & SSIM$\uparrow$ & PSNR$\uparrow$ \\
\midrule
Baseline & 0.6501 & 16.4615 \\
Reg-only & 0.6606 & 16.3403 \\
No-Adversarial & 0.7519 & 17.0020 \\
No-NMI & 0.7618 & 17.2791 \\
Learnable Vector & 0.7548 & 17.5876 \\
Direct Encoding & 0.7597 & 17.0488 \\
Gen-Registration & 0.7956 & 17.4777 \\
Ours & \textbf{0.7926} & \textbf{17.8303} \\
\bottomrule
\end{tabular}
\end{table}
\subsection{Prototype-Driven Virtual Staining}
\label{sec:prototype}

\subsubsection{Style Quantization via Learnable Prototypes}
Our framework extracts the content features of H\&E and IHC images through the explicit content encoder, and the style encoder extracts the staining pattern features of IHC images. While the reference image is visible during training, it is not visible during inference. Therefore, we propose a staining pattern representation based on learnable prototypes. The system maintains $K$ trainable prototypes $\mathcal{P} = \{\mathbf{p}_1,... ,\mathbf{p}_{k}\}$ ($\mathbf{p}_k \in \mathbb{R}^8$) that discretize the continuous style space. Each input style vector $\mathbf{s}$ is quantized through Sinkhorn optimal transport \cite{cuturi2013sinkhorn}.

\begin{equation}
    \mathbf{s}_q = \sum_{k=1}^{K} \alpha_k\mathbf{p}_k, \quad \alpha_k = \text{Sinkhorn}(\mathbf{s},\mathcal{P}, \tau=0.1)
\end{equation}

where the temperature $\tau$ controls assignment sharpness. This soft-quantization approach maintains end-to-end micronutrability while ensuring that all prototypes are utilized in a balanced manner. Notably, these prototypes are not static, but evolve with the training process. Each prototype is updated with momentum based on the mean value of its corresponding staining pattern feature:
\begin{equation}
    \mathbf{p}_k^{(t+1)} \leftarrow \text{Norm}\big(0.9\mathbf{p}_k^{(t)} + 0.1\bar{\mathbf{s}}_k\big)
\end{equation}

where $\bar{\mathbf{s}}_k$ is the mean of features assigned to prototype $k$, and Norm($\cdot$) enforces unit sphere projection. 

\subsubsection{Prototype-Conditioned Image Generation}
The quantized style $\mathbf{s}_q$ drives the staining process through:

\begin{equation}
    \mathbf{\hat{y}} = D(\mathbf{f}_c, \mathbf{s}_q), \quad \mathbf{s}_q = g_\theta(\mathbf{f}_c,\mathcal{P})
\end{equation}

where $g_\theta$ is the prototype aggregator that predicts assignment weights from content features $\mathbf{f}_c$. The decoder $D$ employs AdaIN \cite{huang2017arbitrary} layers to achieve a characteristic blend of content and style:

\begin{equation}
    \text{AdaIN}(\mathbf{f}_c, \mathbf{s}_q) = \gamma\mathbf{s}_q \odot \frac{\mathbf{f}_c-\mu}{\sigma} + \beta\mathbf{s}_q
\end{equation}

\begin{table}[t]
\centering
\caption{Generated Image Quality Evaluation ($Y$ vs $\hat{Y}$)}
\label{tab:generation_quality}
\begin{tabular}{lcccc}
\toprule
Method & SSIM$\uparrow$ & PSNR$\uparrow$ & LPIPS$\downarrow$ & FID$\downarrow$ \\
\midrule
Baseline & 0.3694 & 17.4713 & 0.2913 & 86.1170 \\
Reg-only & 0.3654 & 17.5546 & 0.2755 & 80.1318 \\
No-Adversarial & 0.3687 & 17.7544 & 0.2806 & 85.3821 \\
No-NMI & 0.3665 & 17.5540 & 0.2707 & 82.4281 \\
Learnable Vector & 0.3966 & 18.7112 & 0.3196 & 139.5715 \\
Direct Encoding & 0.3647 & 17.9268 & 0.2564 & 74.1877 \\
Gen-Registration & 0.3845 & 18.3411 & 0.2603 & 82.7541 \\
Ours & \textbf{0.3891} & \textbf{18.5368} & \textbf{0.2501} & \textbf{80.0493} \\
\bottomrule
\end{tabular}
\end{table}

\subsection{\textbf{GapBridge: Deformation-Aware Cross-Modal Registration}}
\label{Deformat}
To address cross-modal misalignment, we propose a dual-path alignment framework that integrates deformation-based registration with content-aware contrastive learning. The framework employs a VoxelMorph network $R$ \cite{balakrishnan2019voxelmorph} to predict a deformation field $\phi = R(X,Y)$, which warps the synthesized image $\hat{Y}$ to align with the target $Y$ through the transformation $\tilde{Y} = W_{\phi}(\hat{Y})$. While this registration step helps correct structural mismatches, we further enhance content consistency between $\hat{Y}$ and the target H\&E images by incorporating a patch-wise contrastive loss $\mathcal{L}_{cont}$, inspired by CUT \cite{cut}. This content-aware contrastive learning mechanism reduces the burden on the generator by explicitly maintaining semantic correspondence between corresponding regions, complementing the geometric alignment achieved through deformation. The combined approach ensures both structural and semantic alignment across modalities.

This joint approach simultaneously addresses three critical challenges in cross-modal image translation:
\begin{itemize}
    \item \textbf{Geometric misalignment}: Differentiable registration corrects structural distortions through $\phi$-based warping
    \item \textbf{Semantic inconsistency}: Contrastive learning maintains content fidelity via $\mathcal{L}_{cont}$
    \item \textbf{Bidirectional registration-migration synergy}: Forward (stain migration $\rightarrow$ cross-modal interference removal) and backward (registration $\rightarrow$ structural consistency $\rightarrow$ accurate stain transfer)
\end{itemize}

\subsubsection{\textbf{Multi-Scale Structural Alignment}}
We propose a hierarchical alignment framework combining three complementary losses to ensure both local and global consistency:

\begin{equation}
    \mathcal{L}_{\text{align}} = \underbrace{\|\tilde{Y} - Y\|_1}_{\text{pixel-level}} + \underbrace{\sum_{l} \|\Phi_l(\tilde{Y}) - \Phi_l(Y)\|_1}_{\text{perceptual}} + \underbrace{\sum_{s} \|P_s(\tilde{Y}) - P_s(Y)\|_1}_{\text{multi-scale}}
\end{equation}

where $\Phi_l$ denotes the $l$-th layer activation of a pretrained VGG network \cite{pati2024accelerating}, and $P_s$ represents Gaussian pyramid decomposition at scale $s$ \cite{liu2022bci}. This unified formulation simultaneously enforces fine-grained intensity matching through L1 loss, maintains high-level semantic consistency via VGG perceptual loss, and preserves structural coherence across spatial scales using pyramid loss.
\subsubsection{\textbf{Transitional deformation suppression}}
While adjacent histological slices share similar anatomical structures, significant variations may still exist due to cutting artifacts, staining differences, or biological variability. Traditional registration methods that enforce strict pixel-level alignment can distort important tissue morphology. Our structure-preserving registration framework addresses this challenge through:

\begin{equation}
\mathcal{L}_{\text{reg}} = \underbrace{-\log\left(\gamma - \text{NMI}(\tilde{Y}, Y)\right)}_{\text{adaptive similarity}} + \underbrace{\lambda\mathcal{R}(\phi)}_{\text{deformation control}}
\end{equation}
By using NMI \cite{guo2024unsupervised} to constrain the similarity between the deformed source image and the target image, along with a smoothness loss \cite{dfmir} to suppress excessive deformation fields, the problem of image distortion and warping can be effectively mitigated.

\subsubsection{Total Registration Objective}

The final objective function for the registration module integrates all constraints:
\begin{equation}
\mathcal{L}_{\text{total}} = \lambda_{\text{cont}} \mathcal{L}_{\text{cont}} + \lambda_{\text{align}} \mathcal{L}_{\text{align}} + \lambda_{\text{reg}} \mathcal{L}_{\text{reg}}
\end{equation}
where $\lambda_{\text{cont}}$, $\lambda_{\text{align}}$ and $\lambda_{\text{reg}}$ control the relative importance of each term. All default to 1 in our experiments.

\subsection{Adversarial Training for Distribution Rendering}
\label{sec:adv}

In traditional adversarial training, it is commonly assumed that the discriminator's role is to determine whether the generated data is real. However, the criteria for this judgment are not explicitly defined. When faced with two unpaired images, should the discriminator focus on assessing structural authenticity or staining pattern authenticity? Based on this observation, we propose that the registration network $R$ assists the generator $G$ in mitigating interference caused by structural misalignment. Simultaneously, this allows the discriminator to concentrate solely on evaluating staining patterns, thereby forming an implicit registration supervision signal.
The adversarial objective consists of:
\begin{multline}
L_{adv} = \min_{G,R}\max_D \Bigl( \mathbb{E}_{Y\sim p_{data}} \left[ \log D(Y) \right] \\
+ \mathbb{E}_{X\sim p_{source}} \left[ \log \left(1 - D(R(G(X))))\right) \right] \Bigr)
\end{multline}
where the expectations are taken over real and generated distributions respectively. \par
The registration supervision emerges through:
\begin{equation}
    \nabla_{\theta_R}L_{adv} = \mathbb{E}_{\tilde{Y}\sim p_G}\left[\frac{\partial L_{adv}}{\partial \tilde{Y}}\cdot\frac{\partial \tilde{Y}}{\partial R}\cdot\frac{\partial R}{\partial \theta_R}\right]
\end{equation}

where $\theta_R$ represents the registration network parameters, and $R$ denotes the registration transformation. This supervision signal is self-driven, meaning the registration network can learn without any ground truth displacement annotations, instead improving through the generator's failures. When the discriminator rejects a synthetic image with structural misalignment, the registration network must collaborate with the generator to produce structurally consistent and realistic staining results. This adversarial cooperation establishes a dynamic game-theoretic framework for geometric reasoning.

\begin{table}[t]
\centering
\caption{Registration Quality Evaluation ($\tilde{Y}$ vs $Y$)}
\label{tab:registration_quality}
\begin{tabular}{lcccc}
\toprule
Method & SSIM$\uparrow$ & PSNR$\uparrow$ & LPIPS$\downarrow$ & FID$\downarrow$ \\
\midrule
Baseline & - & - & - & - \\
Reg-only & 0.4024 & 18.2429 & 0.2592 & 79.8989 \\
No-Adversarial & 0.4469 & 19.1235 & 0.2741 & 89.3116 \\
No-NMI & 0.4083 & 18.3435 & 0.2572 & 80.9087 \\
Learnable Vector & 0.4441 & 19.3734 & 0.3180 & 140.1121 \\
Direct Encoding & 0.4118 & 18.7166 & 0.2367 & 73.2241 \\
Gen-Registration & 0.4877 & 19.8522 & 0.2259 & 82.2434 \\
Ours & \textbf{0.4763} & \textbf{19.8431} & \textbf{0.2315} & \textbf{79.2665} \\
\bottomrule
\end{tabular}
\end{table}

\begin{figure*}[t]
    \centering
    \includegraphics[width=\linewidth]{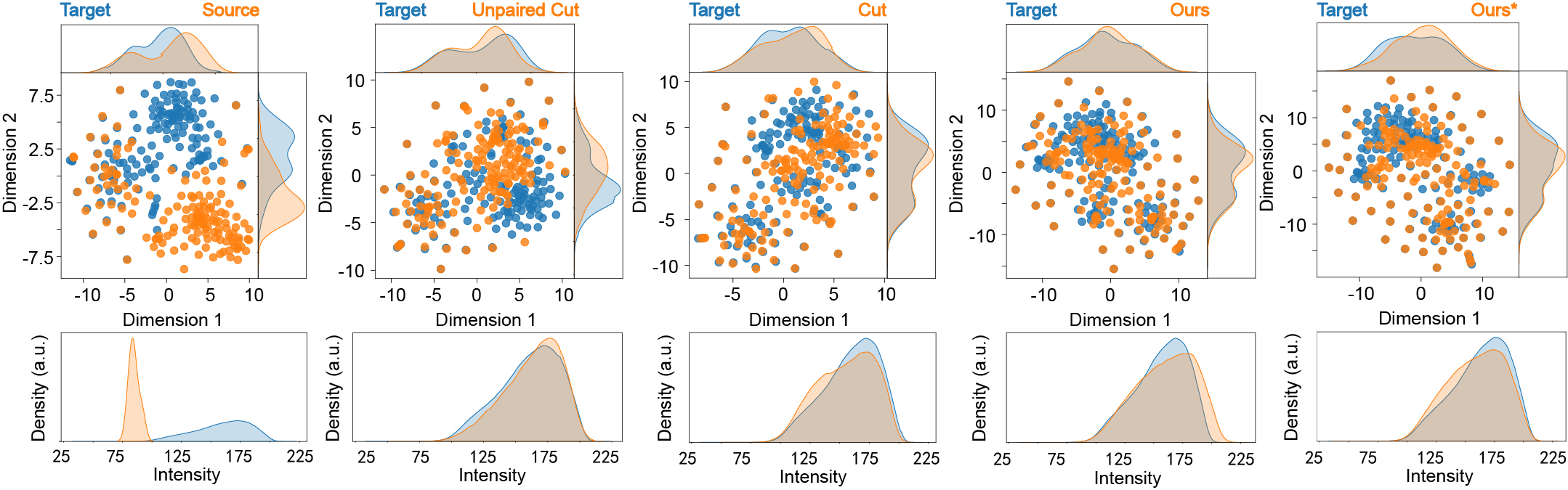}
    \caption{Illustration of the domain offset problem using T-SNE visualization. The yellow and blue dots indicate two images with different domains. From left to right: original H\&E and IHC; image generated by unpaired training setup CUT; image generated by paired training setup CUT; image generated by our method; image after applying the distortion and IHC. the second row shows the distribution of positive cells in the source (generated domain) and target domain samples using color space contrast statistics.}
    \label{fig:story} 
\end{figure*}

\section{Experiments}
\label{sec:experiment}
\subsection{Settings}
\textbf{Dataset.} Since our task is to achieve virtual staining from H\&E to IHC using adjacent tissue slices, we have collected the following datasets:

\begin{enumerate}
    \item \textbf{HIT dataset} \cite{ppt}: It includes 652 image pairs for training and 155 image pairs for testing in the H\&E-CD3 (T-cell marker) dataset, as well as 1614 image pairs for training and 163 image pairs for testing in the H\&E-PAX5 (B-cell marker) dataset.

    \item \textbf{BCI dataset} \cite{liu2022bci}: Collected from breast cancer tissues with a resolution of $1024 \times 1024$, it consists of 3896 H\&E-HER2 pairs for training and 977 pairs for testing.

    \item \textbf{ER dataset from the MIST dataset} \cite{asp}: It includes 4153 pairs for training and 1000 pairs for testing, with a resolution of $1024 \times 1024$.
\end{enumerate}
\textbf{Baseline.} We selected six state-of-the-art baseline methods for a comprehensive comparison, including methods that leverage approximately paired data to tolerate structural inconsistencies, such as FFPE++ \cite{kassab2024ffpe++}, and CUT \cite{cut}, as well as unpaired methods like PyramidPix2Pix \cite{liu2022bci}, ASP \cite{asp}, PSP \cite{psp}, and PPT \cite{ppt}.\par

\textbf{Implementation details.}
We adopt the baseline model provided by the PPT \cite{ppt} framework for training. All images are uniformly resized to \(512 \times 512\), and the model is trained for 200 epochs. All experiments are conducted on an NVIDIA RTX 4090 GPU. For detailed hyperparameter configurations, please refer to our GitHub repository.\par

\textbf{Evaluation metrics.} 
We employ Peak Signal-to-Noise Ratio (PSNR) and Structural Similarity Index Measure (SSIM) to quantify pixel-level similarity between images \cite{wang2004image}. To further assess the quality of generated images, we utilize Fr\'{e}chet Inception Distance (FID) \cite{heusel2017gans} to measure the feature-level distance between image distributions and Learned Perceptual Image Patch Similarity (LPIPS) \cite{zhang2018unreasonable} to evaluate perceptual similarity as perceived by the human visual system. Additionally, we comprehensively analyze the performance of our method by comparing the SSIM and PSNR between the generated images and the original images, as well as the SSIM, PSNR, FID, and LPIPS between the deformed images generated by our method and the target images.
\subsection{Main Results}
\textbf{Qualitative comparison.} Figure \ref{fig:Compare_result} compares the virtual staining performance of our method versus six state-of-the-art methods on four biomarker targets: CD3, PAX5, ER, and HER2. Our method achieves a good trade-off between content preservation and style consistency. Previous methods either overly prioritize content consistency at the expense of realistic staining patterns or excessively align with the target image, leading to distortions in tissue structure. For example, while CUT \cite{cut} and FFPE++ \cite{kassab2024ffpe++} methods utilize contrastive loss to enforce content consistency and learn distribution consistency through discriminators, the lack of an accurate staining model transfer mechanism results in insufficient precision in staining details, particularly in the staining intensity and morphological features of positive cells, which deviate significantly from real staining. The PyramidPix2Pix \cite{liu2022bci} introduces a pixel-level tolerance mechanism, relaxing the strict constraints on structural consistency, thereby generating relatively accurate staining patterns. However, this method underperforms in preserving H\&E image details, leading to the loss of some tissue structures and cellular morphological details, which compromises the completeness and practicality of the staining results. Additionally, ASP \cite{asp}, PSP \cite{psp}, and PPT \cite{ppt} methods attempt to optimize the learning of staining patterns by leveraging the best information provided by adjacent slices while preserving H\&E details. However, these methods heavily rely on data quality, resulting in limited generalization ability and difficulty in adapting to diverse staining scenarios. In contrast, our method accurately preserves H\&E staining details while presenting precise staining patterns. \par

\textbf{Quantitative results.} We quantitatively compared the performance of our proposed method with other SOTA methods on the HIT \cite{ppt}, MIST \cite{asp}, and BCI \cite{liu2022bci} datasets. All compared methods were implemented strictly following the conditions in their original papers or using their open-source code, with the results presented in Table I \ref{tab:evaluation_metrics}. The experiments demonstrate that our method achieves the best performance in terms of content consistency (e.g., SSIM, PSNR) between the generated and original images, as well as staining pattern consistency (e.g., FID) between the generated and target images, fully validating the effectiveness of our virtual staining approach. \par
Specifically, unpaired methods such as CUT \cite{cut} and FFPE++ \cite{kassab2024ffpe++} performed better in terms of content consistency, but the staining pattern consistency was poor due to insufficient supervised signals from neighboring IHC slices. The fully supervised method PyramidPix2Pix \cite{liu2022bci} improves staining through pixel-level optimization, but FID performance is limited by forced alignment of weakly paired images. This problem is particularly prominent in BCI \cite{liu2022bci} datasets containing severely misaligned samples. ASP \cite{asp} and PPT \cite{ppt}, although considering weakly paired features, directly align the generated and target images, ignoring the effect of tissue deformation. PSP \cite{psp} improves performance with the help of pre-trained segmentation networks, but increases computational complexity.\par
Our model demonstrates superior performance advantages across multiple key metrics. Specifically, in terms of content retention performance, the model achieves state-of-the-art results on all datasets except ER protein staining. In terms of FID metrics, our method performs optimally for CD3 (FID=80.0493), PAX5 (FID=81.4670), and ER (fid=32.0808) markers, and only slightly underperforms the PSP \cite{psp} model, which requires an additional cell segmentation step, on the HER2 staining. It is particularly noteworthy that on the CD3 protein, our method achieved the best performance on all evaluation metrics: the FID metric was improved by 6.19\% compared to the suboptimal PSP \cite{psp} model; and the SSIM and PSNR were significantly improved by 21.9\% and 8.32\%, respectively, compared to the baseline method PPT \cite{ppt}.

\subsection{Ablation Study}
Our ablation experiments systematically evaluate the contributions of each key component in the model, specifically designing the following six comparison schemes: (1) Reg-only: remove the style decoupling mechanism of the prototype bootstrap and retain only the alignment module; (2) No-Adversarial: the alignment images are not involved in adversarial training; (3) No-NMI: remove the NMI loss used to constrain the deformation field; (4) Learned Vector: adopting learnable style vectors instead of the decoupling strategy of the prototype library; (5) Direct Encoding: assuming that the target image encoding can be directly obtained at the inference stage; (6) Gen-Registration: inputting the generated images into the deformation network instead of the original H\&E images.\par
The experimental results show that in terms of content retention, all the improved methods can better preserve H\&E image details, with the Gen-Registration method performing closest to the final model, indicating that the participation of the generated image in deformation alignment can effectively enhance the transmission of content information. In terms of coloring pattern consistency, although the Reg-only method performs better in FID metrics, its SSIM and PSNR metrics are poorer due to the lack of an effective decoupling mechanism, which highlights the importance of decoupling the content from the coloring pattern; at the same time, it is difficult to accurately capture the target coloring features in the learnable style vectors, which further verifies the advantage of the prototype-guided strategy. In addition, the alignment network outperforms the unaligned scheme in all evaluation metrics, which confirms that the module not only realizes structural alignment, but also improves the fitting effect of the coloring pattern by reducing the inter-domain differences, and plays a key role as a bridge in the cross-domain transformation. Moreover, if the reference image is available at the time of inference, the in situ staining effect of virtual staining can be further enhanced.\par

\begin{minipage}{\textwidth}
\footnotesize
\end{minipage}
\subsection{Further analysis} 
The experimental results show that the indexes of the aligned images match more closely with the neighboring tissue slices, a finding that is consistent with the results presented in Figure \ref{Fugure1}. To elucidate the principle of the method, we visualized the images generated by different training strategies using the t-SNE technique \cite{van2008visualizing} in Fig. \ref{fig:story}, combined with a statistical comparison of the distribution of the number of positive cells in the color space. The visualization results reveal two key phenomena: first, there is a significant difference between the initial data distribution in the source domain (yellow) and the target domain (blue); second, the distributions in the t-SNE space gradually converge to be close to each other with the improvement of the structural similarity between the input image and the reference image (SSIM) as well as the optimization of the training strategy. This result quantitatively verifies the effectiveness of the alignment network proposed in this study in reducing the inter-domain differences, thus providing strong support for the accurate learning of staining patterns.\par

\section{Conclusion}
Our method demonstrates significant improvements in virtual staining by addressing two critical aspects: preserving detailed information from the original slices and accurately learning the staining patterns. By introducing a deformation network, we effectively reduce structural inconsistencies and minimize the domain gap, leading to enhanced structural alignment and reduced metrics such as LPIPS and FID. The dynamic registration mechanism, driven by adversarial learning, further optimizes the alignment between source and target domains, resulting in more realistic and visually coherent generated images. Additionally, our approach exhibits strong scalability, extending beyond virtual staining to other image transformation tasks with approximate pairing and achieving robust registration capabilities. These advancements underscore the effectiveness and versatility of our method in handling complex image translation challenges.

\section*{Acknowledgements}
This work was supported by the National Key R\&D Program of China (No. 2022YFB4702702).
\bibliographystyle{ACM-Reference-Format}  
\bibliography{sample-base}  

\end{document}